# Computer Vision Applications for Art History: Reflections and paradigms for future research

Amalia Foka
University of Ioannina
Ioannina, Greece
*afoka@uoi.gr*

**One of the contributing factors to the continuing debate among art historians over the use of computational methods in art history research is that they do not consider the core of today's art history research questions. The lack of close collaboration between the two involved research communities makes the definition of contemporary art-historical methods as well-defined computer vision problems extremely difficult. For that purpose, it is devised as a methodology to study articles in art history journals from a computer science perspective. The objective is to identify which image features art historians utilise within their research and describe them in immediate and meaningful terms to the computer vision research community. Finally, some paradigms that could serve as a new starting point for exploring how computer vision applications for art history can address the core of today's art history research are given.**

## 1. INTRODUCTION

Art history research methods are frameworks for viewing artworks, understanding them, and finally concluding into interpretations about their meaning (Hatt & Klonk 2006). Art historians, independently of the research method they utilise, systematically gather and look at artwork images, which either trigger research questions or provide answers for the questions they pose. Hence, it seems inevitable that the computer vision research community would gain interest in exploring computer vision applications for art history.

In Section 2, a short review of computer vision applications for art history is presented. It will be shown that the vast majority of existing computer vision applications for art history are concerned with a minimal range of art-historical research methods. This fact is one of the main contributing factors to the continuing debate over the use of computational methods in art history research, examined in Section 3. On the other hand, as presented in Section 4, the definition of contemporary art-historical methods as computer vision problems is not a straightforward task and requires the close collaboration of computer scientists and art historians, which has not been realised till now. For that purpose, we devise a methodology, presented in Section 5, to study articles in art history journals from a computer science perspective. The objective is first to identify how art historians utilise artwork images within contemporary research methods and, secondly, to describe image features and properties used in art-historical research in immediate and meaningful terms to the computer vision research community. In Section 6, paradigms for future computer vision applications for art history are given. Concluding remarks are presented in Section 7.

## 2. COMPUTER VISION APPLICATIONS FOR ART HISTORY

The following survey's primary purpose is to identify which art-historical methods are explored by current computer vision applications for art history. Hence, the various computer vision methodologies and techniques employed will not be discussed. This type of discussion is included, to some extent, in other surveys present in the literature (Stork 2006; Smith, Stork & Zhang 2009; Brachmann & Redies 2017).

Current computer vision applications for art history are predominately concerned with tasks associated with *Connoisseurship* and *Formalism* (Hatt and Klonk 2006), both established during the 19th and early 20th century. Connoisseurship involves attending to features like brushstrokes, colour, and light as a set of tools for visual analysis. Many computer vision applications exist that replicate the tasks of connoisseurs. For example, to date (e.g., Brown *et al.* 2014), attribute (e.g., Ma et al*.* 2017), and authenticate artworks (e.g., Leonarduzzi, Liu & Wang 2018). Formalism is the study of art by analysing and comparing form and style. Formalists aim to find the style, i.e., the characteristic features, of an individual artist, an epoch, or a culture. Consequently, the tasks of classifying an artwork into a style (e.g., Falomir et al. 2018) and exploring similarities and influences (e.g., Saleh *et al.* 2016) are associated with formalism. Connoisseurship and formalism have in common that they regard only the visual aspects of artworks. This property could attest to the computer vision research community's increased interest in the tasks associated with these methods.

The first stage of the *Iconographical-Iconological* method introduced by Erwin Panofsky in 1920 (Hatt and Klonk 2006) involves identifying natural objects (e.g., humans, animals, houses), their relations, and their expressional qualities (e.g., a pose or gesture). Some applications that attempt to analyse and identify the subject matter of artworks and can be associated with this art-historical method concentrate on identifying the genre (e.g., portrait, landscape) or theme (e.g., flower painting, baptism of Christ) of artworks (e.g., Paul and Malathy 2018). Other applications focus on detecting people in artworks (Westlake, Cai, and Hall 2016) or even recognising specific persons, like Leonardo and other artists (Srinivasan, Rudolph, and Roy-Chowdhury 2015). Some applications can detect a broader set of object categories in artworks (e.g., Crowley and Zisserman 2016). The *Iconographical-Iconological* method was the last and most recent art-historical method we were able to identify in our literature survey of computer vision applications for art history.

Another direction pursued is to provide art historians with tools that facilitate the search and retrieval of artwork images from large databases (Seguin et al*.* 2016; Crissaff et al. 2018). Furthermore, many of the previously mentioned applications for recognising higher-level features, such as objects and people, are utilised to annotate artworks with additional metadata, other than the commonly used (e.g., artist, style, period), in image retrieval systems.

## 3. THE PERSPECTIVE OF ART HISTORIANS

There is a continuing debate over the use of computational methods in art history research, and some prevalent and persistent issues can be identified, explaining why art historians have not yet widely appreciated computer vision applications.

Griselda Pollock (2014) describes current computer vision applications for art history as the kind of research practised in the 19th century or currently in auction houses. Claire Bishop (2017) is also highly critical and states that they exhibit "*a limited awareness of critical debates within art history.*" Gary Hall (2013) regards that most applications consider "*relatively unimportant and unproblematic issues, if not given.*"

Undoubtedly, not all the expressed views on the matter are as harsh. Nevertheless, regardless of how current computer vision applications are appreciated, it is generally acknowledged that they replicate traditional art-historical tasks (Drucker *et al.* 2015) that are now regarded as a primary stage of study and documentation (Schelbert 2017).

As Johanna Drucker (2013) explains, during the 1980s, art history was significantly transformed by the impact of critical theory. Art history was upended by the incursion of semiotics, psychoanalysis, Marxism, feminism, post-colonial theory, and post-structuralism. It changed the way of understanding artworks and the research questions asked. Art history concludes in interpretations with the study of artworks and their accompanying discourse to study the historical context, including the social, political, religious, and geographical conditions in which they were created and gained value. Thus, the emphasis is more on analysing art history's discourses instead of its objects' images (Drucker 2013).

Art historians are not looking for systems that can automate judgments and interpretations. They are convinced that such systems would never be developed, especially if only the images of artworks are considered (Vaughan 2005; Drucker 2013; Klinke 2016). Besides, the perception among art historians is that computer vision research is not yet as advanced as text analysis (Vaughan 2005; Drucker et al*.* 2015) and find it unlikely that it will ever be possible to "*discover the historicity of the work of art in its pixels*" (Rosenberg 2016).

Instead, they are looking for systems that can enhance and augment the art-historical research process, a visual equivalent of word searching (Vaughan 2005) that allows them to discover, chain, and browse artwork images (Rosenberg 2016). Many propose developing an advanced image search and retrieval system as a future direction for computer vision applications for art history (Brandhorst 2013; Drucker 2013; Fletcher 2015; Klinke 2016; Schelbert 2017).

In the previous Section, it has been shown that this direction is already pursued. However, current systems do not enable art historians to ask new and important questions, do research in a new way, or explore relationships that they cannot already (Drucker 2013; Klinke 2016). This seems to be another main reason for existing tools not being widely appreciated and still waiting to be used.

Drucker (2013) envisages a system, which through the identification and analysis of specific features or properties in large corpora, enables to "*situate a work of art within the many networks of cultural relations from which it gains meaning and value*." She provides an example in which pigment analysis of artwork images could be associated with global systems of trade, commerce, and economic value at the material level. Brandhorst (2013) also provides some examples of features and properties of artwork images that would be useful to identify and analyse, although from an iconographical point of view. He also proposes to connect visual and textual content for image search and retrieval.

## 4. THE PERSPECTIVE OF COMPUTER SCIENTISTS

Early art history methods regard only the visual aspects of artworks, and thus, they are easier for computer scientists to comprehend and define as computer vision problems based on low-level image features. On the other hand, more recent art history methods, within the impact of critical theory, study the historical context in which artworks were created and gained value. As a result, defining contemporary art history approaches as well-defined computer vision problems seems an impossible or too complicated task, and the close collaboration of computer scientists and art historians is required.

Nevertheless, it appears that there is a significant gap in understanding between the two involved research communities. Emily L. Spratt and Ahmed Elgammal (2014) have conducted two surveys in 2014 to explore the perceptions of art historians and computer scientists. In the results of these surveys, the vast majority of computer scientists believe that computer science developments have already aided art history. They also firmly believe that the use of artificial intelligence technology in the humanities signals the beginning of a positive paradigm shift. On the other hand, art historians mostly disagree with this statement. However, in the same survey, both parties agree that computer scientists and art historians should work together to benefit from recent developments in artificial intelligence. Still, a large percentage of both parties have never heard of a computer science and art history collaboration.

## 5. WHAT ARE ART HISTORIANS LOOKING FOR IN ARTWORK IMAGES?

As outlined in Section 3, art historians mostly agree that an advanced image search and retrieval system that could enable them to ask new and important research questions would be advantageous. However, it is not clear and straightforward for computer scientists to determine which image features and properties could contribute to the realisation of such a system.

Thus, this paper's central question is, what are art historians looking for in artwork images while analysing the context within which they were created and gained value. Albeit theory and discourse are profoundly important for context analysis, artwork images are also critical in this analysis. However, computer scientists have minimal insight into the image features and properties art historians look for in images (Brandhorst 2013; Drucker 2013).

The utilised methodology was to study articles in art history journals from a computer science perspective. The aim is to compensate for the lack of close collaboration between the two involved research communities. Although the posed question of what art historians are looking for in artwork images might appear too broad or vague, the objective of asking this question is very specific. First, to identify how art historians utilise artwork images for context analysis within the framework of contemporary research methods. Second, describe image features and properties that could trigger or expound art-historical research questions in immediate and meaningful terms to the computer vision research community.

The devised methodology involved studying articles in *The Art Bulletin*, *Art Journal*, and *Art History*. The first two journals are the official publications of the College Art Association (CAA) of America, the preeminent international leadership organisation in the visual arts. The last one is the Association for Art History (AAH) official publication, the subject association for art history in the UK. All three are the most highly ranked journals for art history.

Hence, articles published in these three journals are regarded to represent the leading research in art history adequately. These articles will also represent the various points taken for many art historians' analysis and perspectives. Finally, the study of all articles published in the above three journals in 2017 and 2018 will represent the methods and topics currently prominent in art history research. In conclusion, with the chosen methodology, it will be possible to explore the most persistent and prevalent art historians' requirement;

to consider the core of today's art history research practices and questions.

As expected, our study revealed that there is indeed art history research that does not rely on the analysis of artwork images to conclude in interpretations. However, in many cases, specific image features and properties are utilised during the art history research activities of discovering, chaining, and browsing that will be discussed next.

### 5.1 Image features utilised in current art history research

Following, a categorisation of the image features identified during our study of art history research articles is presented. A brief overview of how these features are utilised in art history research will also be given.

**Detecting *signatures*** to study the artist's presence in relation to the canvas. Guichard (2018) analyses signatures' authorial function and the relationship between the painter and his/her work according to their location on the canvas. Furthermore, particular interest is shown in the analysis of female painters' signatures to study the differences in authorship compared to male painters. Signatures specifically appearing as cartellini, a small slip of paper bearing the artist's signature, are used to study how they were exploited metapictorially, i.e., a picture that discloses something about the nature of painting itself (Ostrow 2017). Signatures detected in Latin American art help explore its relation to European art (Hyman 2017). Finally, signatures that appear on a palette in self-portraits are analysed to reference style and identity (Sohm 2017).

**The *language in which detected text*** was written is utilised to analyse ethnicity or national and cultural identity issues. Signatures in Chinese characters, as well as double signatures in Chinese characters and the Latin alphabet, are utilised to analyse the identity of modern Chinese artists (Clunas 2018). The written text language in public artworks is used to explore the indigenous sovereignty perception (Robinson 2017).

**Recognising *a person's ethnicity,*** either from facial characteristics or clothing, also contributes to analysing national identity issues and colonialism. Coronation durbars in portraits of Maharaja and depictions of British soldiers, along with Indians, facilitate the analysis of British colonialism (Willcock 2017). Similarly, the depiction of Siamese dignitaries facilitates the analysis of French colonialism (Martin 2017). The depiction of Native Americans in contemporary art analyses colonialism's legacies (Morris 2017). Igloliorte (2017) explores the colonial representation of Inuit, the people of the Arctic, previously known as Eskimos. For a similar analysis, detecting a nation's flag, passport, and currency (Cherry 2017) was likewise employed.

**Visual culture *images*** that form the African diasporic culture and identity when found in artworks are used to study how images are transmitted and impact African Americans' perception. Examples of such visual culture images are Aunt Jemima (a brand and trademark of the Quaker Oats Company) and runaway slave advertisements (Elder 2018); the "I AM A MAN" signs held up by striking Memphis sanitation workers in 1968 (Copeland & Thompson 2017); and iconic photographs of the Soweto Uprising in South Africa (Young 2017).

**Landmark *images*** from different periods of time are detected in artworks to examine the political and historical context in which artworks were created. The image of The Tuileries Palace after the fire of 1871 signifies a political event. Its depiction in artworks, as it was before or after the fire, is used to consider the historical context from which these works emerged and the French Third Republic's politics (Wright 2018). Similarly, the depiction of the broken arch of London Bridge speaks of the historical context of the early 19th century where London was viewed as "the new Rome," the first city of a new Empire (Arnold 2017).

**Recognising *biblical figures* and *religious symbols*** are utilised to analyse the changes in perception and religious depiction through time. These include the depiction of Mary and Christ (Brisman 2017; Flora 2017), Moses and Aaron (Merback 2017), nuns (González 2017), worshippers (Rowe 2018), as well as religious decoration (Morrall 2017). A similar analysis is performed by detecting the mythological figures of Cyclops Polyphemus, Perseus, and Andromeda (Jones 2018).

**Recognising the *temporality*** of depicted scenes is used to study the relationship of time with memory and forgetting. The combination of architectural elements from different periods is conceived as an expression of ethnic identity (McWilliam 2017). Rowe (2018) analyses the combination of architecture and clothing of different eras. Finally, a painting that refers to a historical event of the past is utilised to analyse political connotations (Chua 2018).

**Recognising types of *clothing* and their *material*** can also infer temporality as well as the class and status of depicted persons. The cotton dress a woman wears is associated with the trade of textiles (Elder 2018). In the same work, the formal qualities of a dress -unfinished, untailored, and uniform- are associated with the enslaved on plantations in the colonial period. Similarly, Pointon (2017) recognises the importance of linen to class and status. Finally, Flora (2017) analyses the physical connection between Virgin Mary's clothing to Christ's clothing.

**Situating *geographically*** depicted scenes and objects for the analysis of cross-cultural exchanges. The synthesis of landscape scenes originating from widely separated geographical locations is utilised to analyse the impact of colonisation (Hook 2017). The recognition of bamboo leaves creates a geographical-specific setting to analyse modern Chinese artists' identity (Clunas 2018). Recognising patterns on objects originating from different cultures is utilised to explore their meaning in terms of trade relations and others (Rodini 2018).

Although the presented list of image features is neither exhaustive nor complete, it provides enough paradigms on which future computer vision applications for art history could be based on.

## 6. PARADIGMS FOR FUTURE CV APPLICATIONS FOR ART HISTORY

In the following, a discussion is presented on implementing the image features identified in the previous Section and their relation to current research efforts in computer vision.

Previous work on pigment and brushstroke analysis (e.g., Romero et al. 2018) could be redirected to recognise various textiles and other materials. This approach could also contribute to implementing a network that relates materials to global systems of trade, as envisaged by Drucker (2013).

Currently, in computer vision research, there is already an interest in recognising materials and clothing as well as their description by semantic attributes (e.g., Chen, Gallagher and Girod 2012; Liu et al. 2016). Works that emphasise on the recognition of fashion styles (Kiapour et al. 2014) and the evolution of fashion trends (He & McAuley 2016) could enable identifying the period of clothing items. As explained in Section 5, clothing and materials are utilised by art historians to analyse class, status as well as temporality, and hence it would constitute an interesting feature in image search and retrieval systems.

Inferring the temporality of a depicted scene is often used in the analysis of art historians. Temporality can be inferred through the recognition of clothing style but also with the recognition of architectural style. The plausibility of recognising architectural styles of specific epochs is evident from previous computer vision research on the issue *(e.g., Lee et al.* 2015; Shukla et al 2017).

Recognising text and signatures in paintings are also plausible tasks in an already active research direction in computer vision (e.g., Veit *et al.* 2016) and can be applied to detect ethnicity and other identity issues. Identifying a person's ethnicity, gender, and race is also an active research direction based on facial recognition (e.g., Wang et al. 2018). However, a person's ethnicity could also be inferred by the clothing style they wear with approaches similar to those previously described for classifying clothing.

Facial recognition approaches have already been utilised in some computer vision applications for art history to recognise artists and royals (e.g., Srinivasan et al. 2013). These approaches could be extended to recognise mythological, biblical, and religious figures as their depiction in paintings is often used in art-historical research.

Situating geographically depicted scenes by recognising plants can be implemented using various computational botany methods, a highly active research topic (e.g., Wäldchen & Mäder 2018). There is extensive research in computer vision for the geo-localisation of natural environments (Brejcha & Čadík 2017) and the recognition of landmarks (Zheng et al. 2009); however, we regard that these approaches will not be suitable for the tasks as they are involved in art-historical research. Instead, approaches that rely on image similarity with a given input image, a very well-studied topic in computer vision (e.g., Lin et al. 2017), are deemed more appropriate. Image similarity approaches can also recognise input images of landscapes and landmarks as well as visual culture images.

Overall, it has been demonstrated that computer vision research is adequately advanced, and it is feasible to consider the image features identified in Section 5 for future image search and retrieval systems that could augment and enhance art-historical research. Indeed, the discussed image features do not extend to all the possible interests of art historians. However, they can certainly serve as a new starting point for exploring how computer vision applications for art history can address the core of today's art history research practices and questions. Additionally, they can also serve as a new starting point for the closer collaboration of art historians and computer scientists.

Finally, it should be noted that the integration of all the proposed image features in a unified image search and retrieval system could potentially realise the ability for art historians to ask new questions. Nonetheless, consideration should be given to how such concepts and features are defined and how a system is trained to recognise them (Hall 2013). Enabling users' contribution to the definition and training of concepts and features, as in some most recent research paradigms (Teng et al. 2018), could prove beneficial both for the system's ability to generalise and, more importantly, for enabling art historians to fine-tune it according to their research questions. Additionally, recent efforts to connect visual and textual content (e.g., Wu et al. 2017) can also be employed to maximise the richness of utilised features.

Lastly, applying the proposed image search and retrieval system in a large-scale image database that includes both artwork images and visual culture images from multiple sources is crucial for its effectiveness. The copyright of artwork images is a well-known issue (Vaughan 2005), and recent efforts have been directed into utilising the more and more publicly available online images for the materialisation of large-scale databases of artwork images.

## 7. CONCLUSION

This paper has concluded that, to a great extent, art historians envisage an advanced image search and retrieval system that considers the core of today's art history research practices and questions and enables them to ask new and important questions, do research in a new way, or explore relationships that they cannot already.

For that purpose, we have devised as a methodology to study articles in art history journals from a computer science perspective. The aim was first to identify how art historians utilise artwork images for context analysis within the framework of contemporary research methods. Secondly, to describe image features and properties that either trigger or expound art-historical research questions in terms that are immediate and meaningful to the computer vision research community.

This study has revealed some specific image features that are often utilised in current art history research. It has been demonstrated that current computer vision research is adequately advanced, and it is feasible to consider the identified image features for future image search and retrieval systems that could augment and enhance art-historical research. The list of proposed image features we presented is neither exhaustive nor complete. However, the identified image features can undoubtedly serve as a new starting point for exploring how computer vision applications for art history can address the core of today's art history research practices and questions. Additionally, they can also serve as a new starting point for the closer collaboration of art historians and computer scientists.

## 8. REFERENCES

Arnold, D. (2017) Misprisions of London. *Art History*, 40(4), pp. 770–783.

Bishop, C. (2017) Against Digital Art History. *Humanities Futures*.

Brachmann, A. and Redies, C. (2017) Computational and Experimental Approaches to Visual Aesthetics. *Frontiers in computational neuroscience*. Frontiers Media SA, 11, p. 102.

Brandhorst, H. (2013) Aby Warburg's Wildest Dreams Come True? *Visual Resources*. Routledge, 29(1–2), pp. 72–88.

Brejcha, J. and Čadík, M. (2017) State-of-the-art in visual geo-localisation. *Pattern Analysis and Applications*, 20(3), pp. 613–637.

Brisman, S. (2017) The Unassembled Grammar of the Drawing in the Era of Reform. *Art History*, 40(2), pp. 312–335.

Brown, A. D. *et al.* (2014) Can we date an artist's work from catalogue photographs? In *ISPA, 2013 8th Int. Symp,* pp. 558–563.

Chen, H., Gallagher, A. and Girod, B. (2012) Describing Clothing by Semantic Attributes. In Fitzgibbon, A. et al. (eds.) *ECCV 2012*. Springer, pp. 609–623.

Cherry, D. (2017) "Suitcase Aesthetics: The Making of Memory in Diaspora Art in Britain in the Later 1980s," *Art History*, 40(4), pp. 784–807.

Chua, K. (2018) Courbet after Sudjojono. *Art History*, 41(2), pp. 292–317.

Clunas, C. (2018) The Politics of Inscription in Modern Chinese Art. *Art History*, 41(1), pp. 132–153.

Copeland, H. and Thompson, K. (2017) Afrotropes: A User's Guide. *Art Journal*. Routledge, 76(3–4), pp. 7–9.

Crissaff, L. et al. (2018) ARIES: Enabling Visual Exploration and Organization of Art Image Collections. *IEEE Computer Graphics and Applications*, 38(1), pp. 91–108.

Crowley, E. J. and Zisserman, A. (2016) The art of detection. In Hua, G. and Jégou, H. (eds.) *ECCV 2016 Workshops*. Springer, pp. 721–737.

Drucker, J. (2013) Is There a "Digital" Art History? *Visual Resources*. Routledge, 29(1–2), pp. 5–13.

Drucker, J. et al. (2015) Digital art history: the American scene. *Perspective*, 2.

Elder, N. (2018) Lorna Simpson's Fabricated Truths. *Art Journal*. Routledge, 77(1), pp. 30–53.

Falomir, Z. et al. (2018) Categorising paintings in art styles based on qualitative color descriptors, quantitative global features and machine learning (QArt-Learn). *ESWA,* Elsevier, 97, pp. 83–94.

Fletcher, P. (2015) Reflections on digital art history. *CAA Reviews*, 18.

Flora, H. (2017) "Fashioning the Passion: The Poor Clares and the Clothing of Christ," *Art History*, 40(3), pp. 464–495.

González, C. C. (2017) Beyond the Bride of Christ: The Crucified Abbess in Mexico and Spain. *The Art Bulletin*. Routledge, 99(4), pp. 102–132.

Guichard, C. (2018) Signatures, Authorship and Autographie in Eighteenth-Century French Painting. *Art History*, 41(2), pp. 266–291.

Hall, G. (2013) Toward a postdigital humanities. *American Literature*. Duke University Press, 85(4), pp. 781–809.

Hatt, M. and Klonk, C. (2006) *Art history : a critical introduction to its methods*. Manchester University Press.

He, R. and McAuley, J. (2016) Ups and Downs: Modeling the Visual Evolution of Fashion Trends with One-Class Collaborative Filtering. In *Proc. of the 25th Int. Conf. on WWW.*, pp. 507–517.

Hook, G. (2017) Brushes with Infidelity: Truth to Nature in Three Composite Landscapes by Eugene von Guérard. *Art History*, 40(5), pp. 1026–1053.

Hyman, A. M. (2017) Inventing Painting. *The Art Bulletin*. Routledge, 99(2), pp. 102–135.

Igloliorte, H. (2017) Curating Inuit Qaujimajatuqangit. *Art Journal*. Routledge, 76(2), pp. 100–113.

Jones, N. B. (2018) Starting from Places: Continuous Narration and Discontinuous Perspectives in Roman Art. *The Art Bulletin*. Routledge, 100(1), pp. 7–35.

Kiapour, M. H. et al. (2014) Hipster Wars: Discovering Elements of Fashion Styles. In Fleet, D. et al. (eds.) *ECCV 2014*. Springer, pp. 472–488.

Klinke, H. (2016) Big Image Data within the Big Picture of Art History. *Int. Journal for Digital Art History*, (2).

Lee, S. *et al.* (2015) Linking past to present: Discovering style in two centuries of architecture. In *IEEE Int. Conf. on Computational Photography*.

Leonarduzzi, R., Liu, H. and Wang, Y. (2018) Scattering transform and sparse linear classifiers for art authentication. *Signal Processing*. Elsevier, 150, pp. 11–19.

Lin, L. et al. (2017) Cross-Domain Visual Matching via Generalized Similarity Measure and Feature Learning. *IEEE PAMI*, 39(6), pp. 1089–1102.

Liu, Z. et al. (2016) DeepFashion: Powering Robust Clothes Recognition and Retrieval With Rich Annotations.*The IEEE CVPR*.

Ma, D. et al. (2017) From Part to Whole: Who is Behind the Painting? In *Proc. of* ACM MM '17. New York, NY, USA:, pp. 1174–1182.

Martin, M. (2017) History Repeats Itself in Jean-Léon Gérôme's Reception of the Siamese Ambassadors. *The Art Bulletin*. Routledge, 99(1), pp. 97–127.

McWilliam, N. (2017) Towards a New French Renaissance: Memory, Tradition and Cultural Conservatism in France before the First World War. *Art History*, 40(4), pp. 724–743.

Merback, M. B. (2017) 'Between these Two Kingdoms': Exile, Election, and Godly Law in Sebald Beham's Moses and Aaron. *Art History*, 40(2), pp. 286–311.

Morrall, A. (2017) The Family at Table: Protestant Identity, Self-Representation and the Limits of the Visual in Seventeenth-Century Zurich. *Art History*, 40(2), pp. 336–357.

Morris, K. (2017) Crash: Specters of Colonialism in Contemporary Indigenous Art. *Art Journal*. Routledge, 76(2), pp. 70–80.

Ostrow, S. F. (2017) Zurbarán's Cartellini: Presence and the Paragone. *The Art Bulletin*. Routledge, 99(1), pp. 67–96.

Paul, A. and Malathy, C. (2018) An Innovative Approach for Automatic Genre-Based Fine Art Painting Classification. In *Advanced Computational and Communication Paradigms*. Springer, pp. 19–27.

Pointon, M. (2017) Enduring Characteristics and

Unstable Hues: Men in Black in French Painting in the 1860s and 1870s. *Art History*, 40(4), pp. 744–769.

Pollock, G. (2014) Computers can find similarities between paintings – but art history is about so much more. *The Conversation*, 22 Aug. https://theconversation.com/computers-can-find-similarities-between-paintings-but-art-history-is-about-so-much-more-30752

Robinson, D. (2017) Public Writing, Sovereign Reading: Indigenous Language Art in Public Space. *Art Journal*. Routledge, 76(2), pp. 85–99.

Rodini, E. (2018) Mobile Things: On the Origins and the Meanings of Levantine Objects in Early Modern Venice. *Art History*, 41(2), pp. 246–265.

Romero, J., Gómez-Robledo, L. and Nieves, J. (2018) Computational color analysis of paintings for different artists of the XVI and XVII centuries. *Color Research & Application*. Wiley Online Library, 43(3), pp. 296–303.

Rosenberg, R. (2016) Bridging Art History, Computer Science and Cognitive Science. *Zeitschrift für Kunstgeschichte*. Westdeutscher Verlag, 79(3), pp. 305–314.

Rowe, N. (2018) Devotion and Dissent in Late-Medieval Illuminated World Chronicles. *Art History*, 41(1), pp. 12–41.

Saleh, B. *et al.* (2016) Toward automated discovery of artistic influence. *Multimedia Tools and Applications*. Springer, 75(7), pp. 3565–3591.

Schelbert, G. (2017) Art History in the World of Digital Humanities. Aspects of a Difficult Relationship. *Kunsttexte - Renaissance*.

Seguin, B. et al. (2016) Visual Link Retrieval in a Database of Paintings. In Hua, G. and Jégou, H. (eds.) *ECCV 2016*. Springer, pp. 753–767.

Sethi, R. J., Buell, C. A. and Seeley, W. P. (2018) WAIVS: An Intelligent Interface for Visual Stylometry Using Semantic Workflows. In *Proc. of the 23rd Int. Conf. on Intelligent User Interfaces Companion*.

Shukla, P., Rautela, B. and Mittal, A. (2017) A Computer Vision Framework for Automatic Description of Indian Monuments. In *2017 13th Int. Conf. on SITIS*, pp. 116–122.

Smith, B. M., Stork, D. G. and Zhang, L. (2009) Three-dimensional reconstruction from multiple reflected views within a realist painting. *Three-Dimen. Imaging Metrology*, pp. 7239-7239–10.

Sohm, P. (2017) Palettes as Signatures and Encoded Identities in Early-Modern Self-Portraits. *Art History*, 40(5), pp. 994–1025.

Spratt, E. L. and Elgammal, A. M. (2014) The Digital Humanities Unveiled: Perceptions Held by Art Historians and Computer Scientists about Computer Vision Technology. *CoRR*, abs/1411.6.

Srinivasan, R. et al. (2013) Recognising the royals: leveraging computerised face recognition for identifying subjects in ancient artworks. *Proc. of the ACM int. conf. on Multimedia*, pp. 581–584.

Srinivasan, R., Rudolph, C. and Roy-Chowdhury, A. K. (2015) Computerized Face Recognition in Renaissance Portrait Art. *IEEE Signal Processing Magazine*. IEEE, 32(4), pp. 85–94.

Stork, D. G. (2006) Computer Vision, Image Analysis, and Master Art: Part 1. *IEEE MultiMedia*, 13(3), pp. 16–20.

Teng, E., Huang, R. and Iannucci, B. (2018) ClickBAIT-v2: Training an Object Detector in Real-Time. *arXiv preprint arXiv:1803.10358*.

Vaughan, W. (2005) History of art in the digital age: Problems and possibilities. In Gardiner, H., Cashen, T., and Bentkowska-Kafel, A. (eds.) *Digital Art History: A Subject in Transition*.

Veit, A. *et al.* (2016) COCO-Text: Dataset and benchmark for text detection and recognition in natural images. *arXiv preprint arXiv:1601.07140*.

Wäldchen, J. and Mäder, P. (2018) Plant Species Identification Using Computer Vision Techniques. *Archives of Computational Methods in Engineering*, 25(2), pp. 507–543.

Wang, Y. *et al.* (2018) Do They All Look the Same? Deciphering Chinese, Japanese and Koreans by Fine-Grained Deep Learning. In *2018 IEEE Conf. on MIPR*, pp. 39–44.

Westlake, N., Cai, H. and Hall, P. (2016) Detecting People in Artwork with CNNs. In Hua, G. and Jégou, H. (eds.) *ECCV 2016 Workshops*. Springer, pp. 825–841.

Willcock, S. (2017) Composing the Spectacle: Colonial Portraiture and the Coronation Durbars of British India, 1877–1911. *Art History*, 40(1), pp. 132–155.

Wright, A. (2018) On the Origins of Abstraction: {Seurat} and the Screening of History. *Art History*, 41(1), pp. 72–103.

Wu, Q. *et al.* (2017) Visual question answering: A survey of methods and datasets. *Computer Vision and Image Understanding*, 163, pp. 21–40.

Young, A. K. (2017) Visualizing Apartheid Abroad: Gavin Jantjes's Screenprints of the 1970s. *Art Journal*. Routledge, 76(3–4), pp. 10–31.

Zheng, Y. et al. (2009) Tour the world: Building a web-scale landmark recognition engine. In *2009 IEEE CVPR*, pp. 1085–1092.